\documentclass[runningheads]{llncs}
\usepackage[utf8]{inputenc}

\usepackage{physics} 
\usepackage{hyperref}
\usepackage{amsmath,amsfonts,amssymb} 
\usepackage{graphicx}
\usepackage{xcolor}
\usepackage{tabularx}
\usepackage{amsfonts}
\usepackage{array,collcell}

\newcommand\AddLabel[1]{%
  \refstepcounter{equation}% increment equation counter
  (\theequation)% print equation number
  \label{#1}% give the equation a \label
}
\newcolumntype{M}{>{\hfil$\displaystyle}X<{$\hfil}} % mathematics column
\newcolumntype{L}{>{\collectcell\AddLabel}r<{\endcollectcell}}
\usepackage{url}
\usepackage[export]{adjustbox}
\usepackage{subcaption}
\usepackage{graphicx}

\begin{document}
\title{Celestial Machine Learning}
\subtitle{From Data to Mars and Beyond with AI Feynman}
\titlerunning{Celestial Machine Learning}

\author{Zi-Yu Khoo\inst{1} \and Abel Yang\inst{1} \and Jonathan Sze Choong Low\inst{2} \and
Stéphane Bressan\inst{1} }
\authorrunning{Khoo et al.}

\institute{National University of Singapore. 21 Lower Kent Ridge Rd, Singapore 119077
\email{khoozy@comp.nus.edu.sg, phyyja@nus.edu.sg, steph@nus.edu.sg}\\
\and Agency for Science, Technology and Research (A*STAR), Singapore Institute of Manufacturing Technology (SIMTech), Singapore 138634,
Singapore \email{sclow@simtech.a-star.edu.sg}
}
\maketitle              % typeset the header of the contribution
\begin{abstract}
%Title brainstorming: Celestial Machine Learning? Apparatus Doctrina ad Astronomia Nova (machine learning for new astronomy, play on his book titled Astronomia Nova)? 

Can machine learning discover Kepler's first law from data? We emulate Johannes Kepler's discovery of the equations of the orbit of Mars with the Rudolphine tables using AI Feynman, a physics-inspired tool for symbolic regression. 
\end{abstract}

\section{Introduction}
In 2020, Silviu-Marian Udrescu and Max Tegmark introduced \emph{AI Feynman}~\cite{Udrescueaay2631}, a symbolic regression algorithm that could rediscover from data one hundred equations from the \emph{Feynman Lectures on Physics}~\cite{feynmanlecture}. Although the authors motivated their work with the example of Johannes Kepler's successful discovery of the orbital equations of Mars, to our knowledge, they did not report an attempt to rediscover it with their algorithm. We show that AI Feynman can emulate Kepler's discovery of the orbital equation of Mars from the Rudolphine tables. 

The discovery of Kepler's laws of planetary motion illustrates of the process of science, encapsulating the principles of parsimony and physical considerations. Prior to Kepler, early astrologers such as Nicolaus Copernicus and Tycho Brahe hypothesized various models to explain the movement of celestial bodies. Armed with the Prutenic tables, Copernicus modelled the orbit of Mars as heliocentric, with a deferent having two epicycles~\cite{commentariolus_translated}. However, Kepler, assistant to Tycho Brahe, had access to the best available data collected in Europe. In 1627, Kepler compiled Brahe's sightings of Mars into a set of 180 heliocentric data of the position of Mars in the Rudolphine tables. The translation of sightings to the Rudolphine tables only embeds assumptions of heliocentrism and planarity of the orbit. Kepler could have described the motions of Mars as an oval or added additional epicycles to the Copernician model, but instead described it as elliptical in \textit{Astronomia nova} in 1609~\cite{AstronomiaNova}. We use AI Feynman to emulate Kepler's discovery of the elliptical orbital equation of Mars, from the Rudolphine tables. 

% The discovery of Kepler's laws of planetary motion is a prime example of the process of science, encapsulating the principles of parsimony and physical considerations. Prior to Kepler, early astrologers such as Nicolaus Copernicus and Tycho Brahe hypothesized various models to explain the movement of celestial bodies. Armed with the Prutenic tables, Copernicus modelled the orbit of Mars as heliocentric, with a deferent having two epicycles~\cite{commentariolus_translated}. However, Kepler, assistant to Tycho Brahe, had access to the best available data collected using the largest naked-eye instruments in Europe. In 1627, Kepler compiled Brahe's sightings of Mars from 1576 to 1590 into a set of 180 heliocentric data of the position of Mars over half an orbit in the Rudolphine tables. The translation of sightings to the Rudolphine tables only embeds assumptions of heliocentrism and planarity of the orbit. Kepler could have describe the motions of Mars as an oval or added additional epicycles to the Copernician model, but instead described it as elliptical in \textit{Astronomia nova} in 1609~\cite{AstronomiaNova}. 

The Rudolphine tables already embed assumptions regarding planarity and heliocentricity of Mars' orbit. To make further inferences using AI Feynman, we can add biases to the data regarding its physical units. We have four experiments. In the first experiment, AI Feynman is oblivious to biases. In the second and third experiments, AI Feynman is biased through transforming data which are angles and limiting the search space for orbital equations respectively. The fourth experiment combines the biases of the second and third. AI Feynman benefits from these biases, and produces the best result in the fourth experiment. Information regarding the physical units of the data likely also guided Kepler in his discovery of the elliptical orbit of Mars, and in this way, AI Feynman emulates Kepler's discovery from data in the Rudolphine tables. In this paper, we present the design, results and discussion of our experiments with AI Feynman.

\section{Related Work} ~\label{sec:relatedwork}
Finding an equation describing the orbit of Mars is a combinatorial challenge. The task is NP-hard in principle due to its exponentially large search space of equations~\cite{Udrescueaay2631}. To circumvent this, one may use universal function approximators such as multilayer perceptron neural networks~\cite{Hornik_Stinchcombe_White_1989}. Alternatively, symbolic regressions search for a parsimonious and elegant form of the unknown equation. 

There are three main classes of symbolic regression methods~\cite{Makke:2022rnq}: regression-based, expression tree-based and physics- or mathematics-inspired. We use AI Feynman, a machine learning and physics-inspired algorithm~\cite{Udrescueaay2631}.

% \subsection{Regression-based Symbolic Regression Methods}
Regression-based symbolic regression methods~\cite{Makke:2022rnq}, given solutions to the unknown equation, find the coefficients of a fixed basis that minimise the prediction error. As the basis grows, the fit improves, but the functional form of the unknown equation becomes less sparse or parsimonious. Sparse regressions promote sparsity through regularisation, as proposed by Robert Tibshirani~\cite{Tibshirani94regressionshrinkage} who used the $\mathit{l}_1$ norm, thus inventing the Lasso regression.
% Tikhonov~\cite{Tikhonov1963} introduced what is now known as ridge regression, which adds the $\mathit{l}_2$ norm of an equation to its mean squared error. This regularisation can be expressed as Equation~\ref{eq:l2reg}
% \begin{gather}
%     \underset{w \in \mathbb{R}^d}{min} \frac{1}{n} \sum^n_{i=1} V(y_i , \langle \omega,x_i \rangle) +\lambda ||\omega||_2 ~\label{eq:l2reg}
% \end{gather}
% Where $x$ and $y$ are the observed data and $\omega$ is the coefficient to be found through linear regression. 
%Tibshirani~\cite{Tibshirani94regressionshrinkage} introduced a sparse regression via regularisation using the $\mathit{l}_1$ norm of an equation, which he coined LASSO. 
A state-of-the-art sparse symbolic regression approach is the Sparse Identification of Nonlinear Dynamics by Steven Brunton et al. in~\cite{Brunton3932}. It leverages regularisation and identifies equations of motion of a system using a sparse regression over a chosen basis. 
% Shorten: Two state-of-the-art sparse symbolic regression approaches are presented by Kathleen Champion et al. in~\cite{Champion_2019} and Steven Brunton et al. in~\cite{Brunton3932}. Both leverage regularisation. Brunton et al. identified equations of motion of a system using a sparse regression over a chosen basis called Sparse Identification of Nonlinear Dynamics. Champion et al. discovered a low dimensional space where equatiosn could be sparsely represented to apply the Sparse Identification of Nonlinear Dynamics.

% \subsection{Expression Tree-based Symbolic Regression Methods}

Committing to a basis limits the applicability of regression-based methods. Expression tree-based symbolic regression methods based on genetic programming~\cite{Makke:2022rnq} can instead discover the form and coefficients of the unknown equation. 

Seminal work by John Koza et al.~\cite{Koza92} represented each approximation of an unknown equation as a genetic programme with a tree-like data structure, with traits (or nodes in the tree) representing functions or operations, and variables representing real numbers. The fitness of each genetic programme is its prediction error. Fitter genetic programmes undergo a set of transition rules comprising selection, crossover and mutation to iteratively find the optimal equation form. 

% Shortened: Genetic programming was first applied to symbolic regression in the seminal work by John Koza et al.~\cite{Koza92}. Each approximation of an unknown equation is a genetic programme represented by a tree-like data structure, the traits (or nodes in the tree) of which are functions, operations, and variables representing real numbers~\cite{Koza92}. The fitness of each genetic programme is calculated between the predicted output and the given solutions to the unknown equation. Fitter genetic programmes are selected to undergo a set of transition rules comprising selection, crossover and mutation to find the optimal structure over many iterations. 

Genetic programmes may greedily mimic nuances of the unknown equation~\cite{Smits2005}, limiting generalisability. 
David Goldberg~\cite{Goldberg1989} used Pareto optimisation to balance the objectives of fit and parsimony in symbolic regression. In each iteration, the fittest genetic programmes lie on the non-dominated Pareto-frontier. State-of-the-art symbolic regression using genetic programming include \verb|Eureqa| by Michael Schmidt and Hod Lipson~\cite{Schmidt81} and \verb|PySR| by Miles Cranmer~\cite{pysr}. 

% Shortened: Multiple genetic programmes may describe the unknown equation well. Some genetic programmes may even greedily mimic each nuance between inputs and outputs of the unknown equation~\cite{Smits2005}, limiting the generalisability of the selected genetic programmes. 
% To overcome this, David Goldberg~\cite{Goldberg1989} suggested using Pareto optimisation. In Pareto optimisation, a Pareto-frontier balances the two objectives of fitting the genetic programme to the output of the unknown equation and reducing the complexity of the genetic programme. In each iteration, the non-dominated Pareto-frontier is identified and used to select genetic programmes to undergo transitions for the next iteration. 

% \subsection{Physics-Inspired or Mathematics-Inspired Symbolic Regression Methods}
Expression tree-based methods do not guarantee that more accurate approximations of an equation are symbolically closer to the truth. If an expression tree-based method finds a reasonably accurate equation with wrong functional form, it risks getting stuck near a local optimum~\cite{Udrescueaay2631}. Functions of practical interest in physics exhibit simplifying properties such as symmetry or separability~\cite{Udrescueaay2631}. Physics-inspired symbolic regression methods leverage these simplifying properties to guarantee taking a step in the right direction. Udrescu and Tegmark~\cite{Udrescueaay2631} use a neural network to test data describing the unknown equation for simplifying properties and recursively break the unknown equation into simpler unknown equations with fewer variables to solve~\cite{Udrescueaay2631}. Each simpler unknown equation can be solved by regression with a basis set of non-linear functions. AI Feynman then outputs a sequence of increasingly complex equations that provide progressively better accuracy, along a Pareto front, based on work by Goldberg~\cite{Goldberg1989} and Guido Smits~\cite{Smits2005}. We use AI Feynman to rediscover the orbital equation of Mars.

\section{Methodology} ~\label{sec:methodology}

In publishing the Rudolphine tables, Kepler had already assumed planarity and heliocentricity of the orbit of Mars. We experiment if AI Feynman performs better with biases, based on our knowledge of the physical units of the data. 

Informing a learning algorithm of physics amounts to introducing appropriate biases that can steer the learning process towards identifying physically consistent solutions according to George Karnadiakis~\cite{Karniadakis2021}. Karniadakis identifies three types of bias: observational, inductive and learning biases.
Observational biases are introduced directly through data that embody the underlying physics or carefully crafted data augmentation procedures. Training a machine learning model on such data allows it to learn an output that reflects the physical structure of the data. Inductive biases correspond to prior assumptions incorporated by tailored interventions to a machine learning model architecture, so predictions are guaranteed to satisfy a set of given physical laws. Learning biases can be introduced by appropriate choice of loss functions, constraints and inference algorithms that can modulate the training phase of an machine learning model to
explicitly favour convergence towards solutions that adhere to the underlying physics. We consider the introduction of observational and inductive biases.

With knowledge that data is known to be an angle, only trigonometric functions can transform it. We introduce an observational bias by applying the sine and cosine functions to inputs of the unknown equation that are known to be angles. The resulting numerical values, hence embodying the underlying periodicity of the data, are input to AI Feynman. The observational bias guides AI Feynman in finding an equation that reflects the periodic nature of Mars' orbit. 

With knowledge that data are physical quantities, they cannot be transformed by exponential and logarithmic functions, as these only transform dimensionless quantities. We introduce an inductive bias by eliminating candidate functions. For each simpler recursive unknown equation AI Feynman has to solve, it transforms equations in the current Pareto front by one of several non-linear functions. These non-linear functions include exponential, logarithmic, trigonometric, polynomial and radical functions. %Introducing an inductive bias prevents AI Feynman from transforming equations in the current Pareto front by selected non-linear functions. The inductive bias strictly limits the search space of AI Feynman, as it finds an equation using only non-selected functions. 
The inductive bias limits the search space to trigonometric, polynomial and radical functions.

We conduct four experiments corresponding to four possible combinations of observational and inductive biases with the AI-Feynman algorithm. Experiment 1 does not use any bias. Experiment 2 and 3 only use observational and inductive bias respectively. Experiment 4 uses both observational and inductive bias.

While AI Feynman explores the Pareto front, Kepler may have instead made use of thought experiments to hypothesize an elliptical orbit. Fitting data from the Rudolphine tables to the equation of an ellipse using non-linear least squares returns the coefficients representing the eccentricity and semi-major axis. These are $0.0926$ and $1.5235$ respectively. For reference, the National Aeronautics and Space Administration suggest an $0.0934$ and $1.5237$ respectively~\cite{NASA2020MarsFacts}. 

%We report the experimental setup and results in Section~\ref{sec:evaluation}.

% We conduct three experiments that vary in their physics-informed machine learning augmentations to the AI-Feynman algorithm. Experiment 1 makes use of an observational bias and an inductive bias that excludes all functions except the polynomial and inverse functions. Experiment 2 makes use of only an observational bias. Experiment 3 does not make use of any bias. We report the experimental setup and results in Section~\ref{sec:evaluation}.

\section{Performance Evaluation} ~\label{sec:evaluation}
\begin{figure}
    \centering
    \includegraphics[width = 0.7\textwidth]{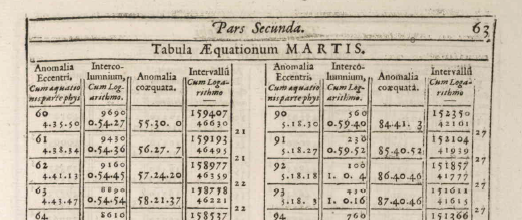}
    \caption{The four columns of data provided in the Rudolphine Tables.}
    \label{fig:pg63mars}
\end{figure}

In the Rudolphine tables~\cite{rudolphine_tables}, the table titled \textit{Tabula Aequationum MARTIS}, or Table of Corrections for Mars, contains four columns of data: \textit{Anomalia eccentri}, \textit{Intercolumnium}, \textit{Anomalia coaequata}, and \textit{Intervallu}. These columns represent the eccentric anomaly, an interpolating factor, the coequated or true anomaly, and the distance between the Sun and Mars respectively. The full Rudolphine tables (snippet found in Figure~\ref{fig:pg63mars}) was digitised for this experiment.

We apply AI Feynman to the data of \textit{Intervallu} and \textit{Anomalia coaequata} to recover the equation of orbit for Mars. \textit{Anomalia coaequata} is an angle in degrees minutes seconds, which we convert to decimal degrees. \textit{Intervallu} is the distance between the Sun and Mars, which we scale from a magnitude of $E+05$ to $E+00$. The code for AI Feynman, with minor modifications to embed observational and inductive biases, is at \url{https://github.com/zykhoo/AI-Feynman}.

We compare equations along the AI Feynman Pareto frontier with the orbital equation of Mars in Equation~\ref{eqn:mars_eqn}. $r$ is the \textit{Intervallu} and $\theta$ is the \textit{Anomalia coaequata}. $\epsilon$ is the eccentricity of the ellipse, and $a$ is the semi-major axis. 
\small{
\begin{gather}
    r = \frac{a}{1+\epsilon \times cos\theta} \label{eqn:mars_eqn} \tag{0}
\end{gather}
}%
\normalsize
We also present the mean description length loss\cite{Udrescueaay2631,wutailin_thesis} (DL) computed between each predicted and true \textit{Intervallu}. It minimises the geometric mean instead of the arithmetic mean, which encourages improving already well-fit points~\cite{wutailin_thesis}. 

For Experiments 1 and 3, the inputs to AI Feynman are $\theta$ and $r$. For Experiments 2 and 4, the inputs to AI Feynman are $\cos(\theta)$, $\sin(\theta)$ and $r$. We traverse the equations along the Pareto frontier returned by AI Feynman in increasing goodness of fit and increasing complexity (or equivalently decreasing parsimony) and present them. The results of Experiments 1, 2, 3 and 4 are presented in Tables~\ref{tab:exp1}, ~\ref{tab:exp2}, ~\ref{tab:exp3} and \ref{tab:exp4} respectively. We omit results independent of $\theta$.

\begingroup
\begin{subequations}
\begin{table}[h]
    \tiny
    \centering
    \renewcommand{\arraystretch}{1.25} % Default value: 1
    \begin{tabularx}\textwidth{|@{}|L|M|c|@{}|}
    \hline
    \multicolumn{1}{|c|}{\textbf{Eqn No.}} & \multicolumn{1}{|c|}{\textbf{Equation}} & \textbf{DL}    \\ 
    \hline
    % 1 & r=1/(666.000000000000+(\sin(\pi))^{-1})& 30.604 \\
    % 2 & r=1.5 & 26.186 \\
    % 3 & r=\pi/2 & 26.186 \\
    % 4 & r=1.65306122448980 & 26.178 \\
    5 & r=\frac{4}{3} - 0.09\times \theta^2 & 24.976 \\
    6 & r=(2.78 - 0.26\times \theta^2)^0.5 & 24.926 \\
    7 & r=\arccos{(-0.02\times \theta^3 + 0.09\times \theta^2 - 0.1)}& 23.577 \\
    8 & r=\frac{1}{-0.01\times \theta^3 + 0.04\times \theta^2 + 0.6} & 22.515 \\
    9 & r=(0.01\times \theta^3 - 0.04\times \theta^2 + 1.29)^2 & 22.273 \\
    10 & r=\arccos{(-0.02\times \theta^3 + 0.09\times \theta^2 + 0.01\times \theta - 0.1)} & 21.356 \\
    11 & r = \log(0.09\times \theta^3 - 0.38\times \theta^2 - \frac{1}{9}\times \theta + 5.3) & 20.841 \\
    12 & r=0.02\times \theta^3 - 0.09\times \theta^2 - 0.01\times \theta + 1.67 & 20.238 \\
    \hline
    \end{tabularx}
    \caption{Results of Experiment 1, which took 748 seconds.}
    \label{tab:exp1}
\end{table}
\end{subequations}
\endgroup

% For Experiment 2, the inputs to AI Feynman are $\cos(\theta)$, $\sin(\theta)$ and $r$. We traverse the equations along the Pareto frontier returned by AI Feynman in increasing goodness of fit and increasing complexity (or equivalently decreasing parsimony) and present them in Table~\ref{tab:exp2}. 

\begingroup
\begin{subequations}
\begin{table}[h]
    \tiny
    \centering
    \renewcommand{\arraystretch}{1.3} % Default value: 1
    \begin{tabularx}\textwidth{|@{}|L|M|c|@{}|}
    \hline
    \multicolumn{1}{|c|}{\textbf{Eqn No.}} & \multicolumn{1}{|c|}{\textbf{Equation}} & \textbf{MSE}    \\ 
    \hline
    % 12a & r = 1.5 & 26.186 \\
    % 13 & r = 1.42857142857143 & 26.182 \\
    14 & r = \log{\cos\theta + 5} & 26.006 \\
    15 & r = \frac{1}{7} \times \cos\theta + 1.5 & 24.053 \\
    eqn:exp2exp1 & r = 1.5\times \exp{0.1\times \cos\theta} & 23.512 \\
    eqn:exp2cor1 & \mathbf{r = \frac{1}{\frac{2}{3} - 0.0556244812357114\times \cos\theta} } & 22.857 \\ 
    eqn:exp2exp2 & r = 1.5119670200057298\times \exp{(0.1\times \cos\theta)} & 22.457 \\
    20 & r = 1.510965630582+(\cos\theta/(\sin\theta+6)) & 21.070 \\
    eqn:exp2exp3 & r = 1.51366746425629\times exp(0.0931480601429939\times \cos\theta) & 20.762 \\
    eqn:exp2cor2 & \mathbf{r = \frac{1}{0.662428796291351 - 0.0612906403839588\times \cos\theta}} & 19.781 \\
    eqn:exp2cor3 & \mathbf{r = (0.662428796291351 - 0.0612906403839588\times \cos\theta)^{-1.00133872032166}} & 12.211 \\
    \hline
    \end{tabularx}
    \caption{Results of Experiment 2, which took 1451 seconds.}
    \label{tab:exp2}
\end{table}
\end{subequations}
\endgroup

% For Experiment 3, the inputs to AI Feynman are $\theta$ and $r$. We traverse the equations along the Pareto frontier returned by AI Feynman in increasing goodness of fit and increasing complexity (or equivalently decreasing parsimony) and present them in Table~\ref{tab:exp3}. 

\begingroup
\begin{subequations}
\begin{table}[h]
    \tiny
    \renewcommand{\arraystretch}{1.25} % Default value: 1
    \centering
    \begin{tabularx}\textwidth{|@{}|L|M|c|@{}|}
    \hline
    \multicolumn{1}{|c|}{\textbf{Eqn No.}} & \multicolumn{1}{|c|}{\textbf{Equation}} & \textbf{MSE}    \\ 
    \hline
    % 26 & r=\arcsin(-666.000000000000\times \sin(\pi)) & 30.604 \\
    % 27 & r=1.5 & 26.186 \\
    % 28 & r=\pi/2 & 26.186 \\
    % 29 & r=1.65306122448980 & 26.178 \\
    30 & r=\frac{4}{3} - 0.09\times \theta^2 & 24.976 \\
    31 & r=(2.78 - 0.25\times \theta^2)^0.5 & 24.842 \\
    32 & r=\arccos{(-0.02\times \theta^3 + 0.09\times \theta^2 - 0.1)} & 23.577 \\
    33 & r=\frac{1}{-0.01\times \theta^3 + 0.04\times \theta^2 + 0.6} & 22.515 \\
    34 & r=(0.01\times \theta^3 - 0.04\times \theta^2 + 1.29)^2 & 22.273 \\
    35 & r=\arccos{(-0.02\times \theta^3 + 0.09\times \theta^2 + 0.01\times \theta - 0.1)} & 21.356 \\
    37 & r=0.02\times \theta^3 - 0.09\times \theta^2 - 0.01\times \theta + 1.67 & 20.238 \\
    \hline
    \end{tabularx}
    \caption{Results of Experiment 3, which took 621 seconds.
}
    \label{tab:exp3}
\end{table}
\end{subequations}
\endgroup

% For Experiment 4, the inputs to AI Feynman are $\cos(\theta)$, $\sin(\theta)$ and $r$. We traverse the equations along the Pareto frontier returned by AI Feynman in increasing goodness of fit and increasing complexity (or equivalently decreasing parsimony) and present them in Table~\ref{tab:exp4}. 
% {\scriptsize
% \begin{gather}
% y = \frac{1}{7}\times x_0 + 1.5 \\
% y = 0.142857142857143\times x_0 + 1.51698136329651 \\
% y = 0.141145795583725\times x_0 + 1.51698136329651 \\
% y = 0.501939036628\times (x_0\times (x_0 + 1) + 1)^0.25 + 1.003878073256 \\
% y = 0.501939036628\times ((sqrt(sqrt(((x_0\times (x_0+1))+1)))+1)+1) \\
% y = 0.140863761305809\times x_0 - 0.0146047221496701\times x_1 + 1.52623343467712 
% \end{gather}
% }%
\begingroup
\begin{subequations}
\begin{table}[h]
    \tiny
    \renewcommand{\arraystretch}{1.25} % Default value: 1
    \centering
    \begin{tabularx}\textwidth{|@{}|L|M|c|@{}|}
    \hline
    \multicolumn{1}{|c|}{\textbf{Eqn No.}} & \multicolumn{1}{|c|}{\textbf{Equation}} & \textbf{MSE}    \\ 
    \hline
    % 38 & r=1.5 & 26.186 \\
    40 & r=\frac{1}{7}\times \cos\theta + 1.5 & 24.053 \\
    eqn:exp4x0x11 & r=\cos\theta/(\sin\theta + 6) + 1.5 & 23.617 \\
    41 & r=\arccos(0.0420224035468255 - 0.142857142857143\times \cos\theta) & 23.392 \\
    eqn:exp4cor1 & \mathbf{r=\frac{1}{\frac{2}{3} -  0.0566732120453772\times \cos\theta}} & 22.575 \\
    eqn:exp4x0x12 & r=1.511006320056+(\cos\theta/(\sin\theta+6)) & 21.089 \\
    42 & r = \tan(0.0425049090340329\times \cos\theta + 0.986141372332807) & 20.057 \\
    43 & r = \tan(0.0427569970488548\times \cos\theta + 0.98658412694931) & 20.021 \\
    eqn:exp4cor2 & \mathbf{r=\frac{1}{0.662420213222504 - 0.0612917765974998\times \cos\theta}} & 19.747 \\
    eqn:exp4cor3 & \mathbf{r=(0.662420213222504 - 0.0612917765974998\times \cos\theta)^{-1.00130701065063}} & 12.208 \\
    \hline
    \end{tabularx}
    \caption{Results of Experiment 4, which took 1184 seconds.}
    \label{tab:exp4}
\end{table}
\end{subequations}
\endgroup

Experiment 1 does not use any bias. The search space is big and AI Feynman does not find an equation form that matches the orbital equation of Mars along its Pareto frontier. In Experiment 2, AI Feynman makes use of an observational bias. As the input data to AI Feynman embodies the underlying periodicity of the data, AI Feynman can use this information to guide its search for an equation that reflects the periodic structure of Mars' orbit. Therefore, three out of nine equations along the Pareto front have an equation form that matches the true orbit of Mars. These are Equations~\ref{eqn:exp2cor1},~\ref{eqn:exp2cor2} and~\ref{eqn:exp2cor3}. In Experiment 3, AI Feynman makes use of an inductive bias. While the search space for AI Feynman is smaller, an inductive bias does not guide its search for an equation that reflects the periodic structure of Mars' orbit. Therefore AI Feynman does not find an equation form that matches the orbital equation of Mars along its Pareto frontier. In Experiment 4, AI Feynman makes use of both an observational and an inductive bias. AI Feynman makes use of the observational bias to guide its search for an equation that reflects the periodic structure of Mars' orbit. It also makes use of the inductive bias to limit the search space, resulting in fewer equations along the Pareto front. Therefore, three out of ten equations along the Pareto front have an equation form that matches the true orbit of Mars. These are Equations~\ref{eqn:exp4cor1},~\ref{eqn:exp4cor2} and~\ref{eqn:exp4cor3}.

Experiments 1 and 2 highlight the importance of an observational bias in guiding AI Feynman. In Experiment 1, none of the equations along the Pareto front match the orbital equation for Mars, compared to three out of eleven equations along the Pareto front for Experiment 2. However, as the observational bias doubles the number of inputs to AI Feynman, it takes approximately twice as long to run. This is because AI Feynman recurses through each input. The depth of the recursion is doubled when the number of inputs is doubled.  

Experiments 2 and 4 highlight the importance of an inductive bias in limiting the search space for AI Feynman. In Experiment 2, we can observe many equations have one of two common forms. Three of them utilise an exponential function applied to $\cos\theta$ (Equations~\ref{eqn:exp2exp1}, \ref{eqn:exp2exp2} and \ref{eqn:exp2exp3}). Another three utilise an inverse function applied to $\cos\theta$ (Equations~\ref{eqn:exp2cor1},~\ref{eqn:exp2cor2} and~\ref{eqn:exp2cor3}), which matches the true orbit of Mars. However, in Experiment 4, the equation form with an inverse function applied to $\cos\theta$ in (Equations~\ref{eqn:exp4cor1},~\ref{eqn:exp4cor2} and~\ref{eqn:exp4cor3}) matches the true obit of Mars, and is also the most prevalent. Therefore, AI Feynman, augmented with both an observational and inductive bias, is best able to rediscover Kepler's first law for the orbit of Mars. The inductive bias also reduces the time taken for AI Feynman to run. This is because the search space for AI Feynman is limited, and time is saved from having to search a smaller search space. 

We also observe that Equations~\ref{eqn:exp2cor2},~\ref{eqn:exp2cor3}, \ref{eqn:exp4cor2}, and \ref{eqn:exp4cor3}, with forms that match the true orbit of Mars, have the lowest mean description length loss of less than $20$.

Lastly, we observe that Equations~\ref{eqn:exp2cor2} and \ref{eqn:exp4cor2} suggest $a=1.52$ and $\epsilon=0.0925$ similar to the values suggested in the Rudolphine tables.

\section{Conclusion} ~\label{sec:conclusion}
We have successfully shown that AI Feynman can rediscover from the Rudolphine Tables the equation of Kepler's first law for the planet Mars, given information regarding physical quantities of the data in the form of observational and inductive biases. The discovery of physical laws is a bi-optimisation problem of parsimony and accuracy, that can be guided by physical units of the data available. AI Feynman is able to emulate Kepler's discovery of the orbital equation of Mars because it implements an optimisation of both parsimony and accuracy, and can be guided by information regarding the physical units of the data. % (value judgement: emulation because AI Feynman is implementing a combination of best fit and parsimony) conclusion: that the discovery in physics is an bi-optimisation problem of accuracy and parsimony, occam razor. 

% we show that Kepler's symbolic regression of these observations yielding an ellipse is the optimal in terms of fit and parsimony. 

% We show that AI Feynman can repeat the work of Kepler
% can say later that AI Feynman can emulate the best and most parsiomonious fit 

% our qualitative claim: AIF emulates kepler's thinking because it optimises both accuracy and parsimony and this is probably a generic scientific methodology (as suggested by Occams razor).  

As future work, we are looking into how AI Feynman can repeat this discovery process directly from sightings of Mars and the Sun from Earth. We use a modern reproduction of these sightings from the National Aeronautics and Space Administration's Horizons system. This challenges AI Feynman to perform a change from the geocentric to heliocentric reference frame, and we are investigating how this change can be incorporated within its algorithm. Lastly, we are experimenting with the planet Mercury, which has a precessing orbit.

%\section*{Acknowledgement} This research is partially supported by the Agency of Science, Technology and Research (A*STAR), by the National Research Foundation, Prime Minister’s Office, Singapore, under its Campus for Research Excellence and Technological Enterprise (CREATE) programme as part of the programme DesCartes, and by the Ministry of Education, Singapore, under its Academic Research Fund Tier 2 grant call (Award ref:  MOE-T2EP50120-0019). Any opinions, findings and conclusions or recommendations expressed in this material are those of the authors and do not reflect the views of the National Research Foundation or of the Ministry of Education, Singapore. 

\bibliographystyle{splncs04}
\bibliography{cite}

\section*{Appendix}
\begin{table}[]
    \centering
    \begin{tabular}{|c|c|c|}
    \hline
    Eqn No. & Eqn & MSE \\
    \hline
    1 & $asin(-666.000000000000\times ((sin(pi)+1)-1))$ & 2.33  \\
    2 & $1.50000000000000$ & 0.0106 \\
    3 & $pi/2$ & 0.0123 \\
    4 & $1.65306122448980$ & 0.0268 \\
    5 & $1.66666666666667 - 0.09\times x_0^2$ & 0.048 \\
    6 & $(2.78 - 0.26\times x_0^2)^0.5$ & 0.0767 \\
    7 & $acos(-0.02\times x_0^3 + 0.09\times x_0^2 - 0.1)$ & 0.000169 \\
    8 & $1/(-0.01\times x_0^3 + 0.04\times x_0^2 + 0.6)$ & 0.000706 \\
    9 & $(0.01\times x_0^3 - 0.04\times x_0^2 + 1.29)^2$ & 0.000458 \\
    10 & $acos(-0.02\times x_0^3 + 0.09\times x_0^2 + 0.01\times x_0 - 0.1)$ & 4.7e-05 \\
    11 & $log(0.09\times x_0^3 - 0.38\times x_0^2 - 0.111111111111111\times x_0 + 5.3)$ & 1.07e-05 \\
    12 & $0.02\times x_0^3 - 0.09\times x_0^2 - 0.01\times x_0 + 1.67$ & 4.41e-05 \\
    \hline
    \end{tabular}
    \caption{Experiment 1 Results}
    \label{tab:my_label1}
\end{table}
\begin{figure}[h]
    \centering
    \includegraphics[width = 0.7\textwidth]{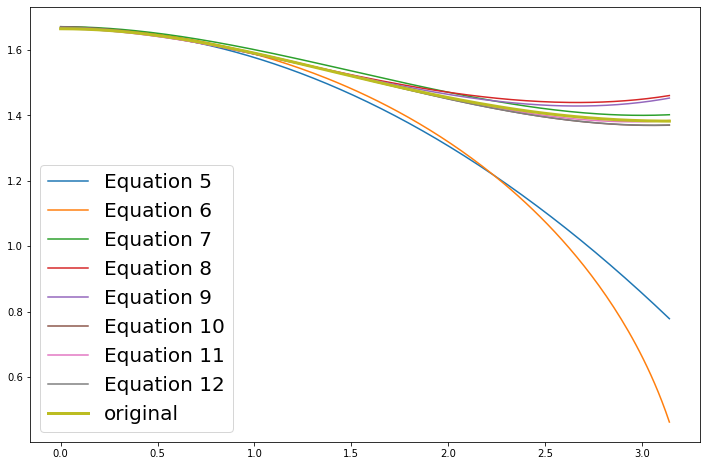}
    \caption{A plot of the results of Experiment 1. The equations correspond to those presented in Table~\ref{tab:exp1}. The y-axis represents the \textit{Intervallu} and x-axis represents the \textit{Anomalia coaequata}. The true values for the \textit{Intervallu} from the Rudolphine tables are also plotted and labeled "original".}
    \label{fig:exp1}
\end{figure}
\begin{table}[]
    \centering
    \begin{tabular}{|c|c|c|}
    \hline
    Eqn No. & Eqn & MSE \\
    \hline
    1 & $1.50000000000000$ & 0.0106 \\
    2 & $log(x_0 + 5)$ & 0.0092 \\
    3 & $0.142857142857143\times x_0 + 1.5$ & 0.000309 \\
    4 & $1.5\times exp(0.1\times x_0)$ & 0.00021 \\
    5 & $acos(0.0418258837357514 - 0.142857142857143\times x_0)$ & 0.000166 \\
    6 & $1/(0.666666666666667 - 0.0571196591636008\times x_0)$ & 0.000212 \\
    7 & $1.5082674607662332\times exp(0.1\times x_0)$ & 7.44e-05 \\
    8 & $1.510965630582+(x_0/((((((x_1+1)+1)+1)+1)+1)+1))$ & 0.000179 \\
    9 & $1.51366746425629\times exp(0.0931480601429939\times x_0)$ & 5.39e-06 \\
    10 & $tan(0.0427570976316929\times x_0 + 0.986583888530731)$ & 1.98e-06 \\
    11 & $tan(0.0428397443727006\times x_0 + 0.986126406475687)$ & 4.21e-06 \\
    12 & $1/(0.662416338920593 - 0.0612923018634319\times x_0)$ & 7.26e-07 \\
    13 & $(0.662416338920593 - 0.0612923018634319\times x_0)^(-1.0012925863266)$ & 6.01e-10 \\
    \hline
    \end{tabular}
    \caption{Experiment 2 Results}
    \label{tab:my_label2}
\end{table}

\begin{figure}[h]
    \centering
    \includegraphics[width = 0.7\textwidth]{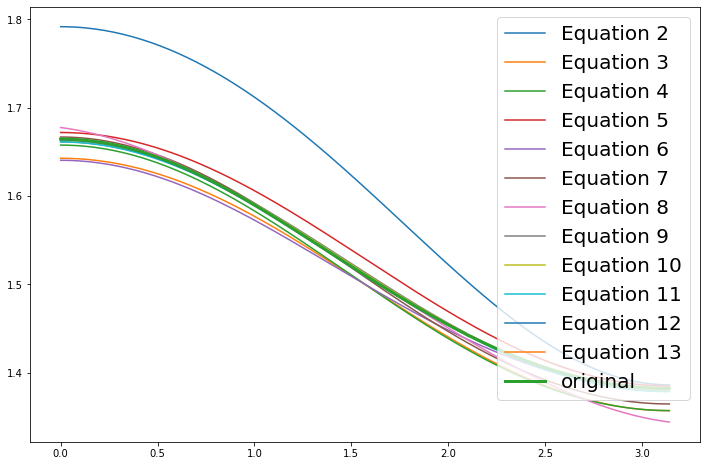}
    \caption{A plot of the results of Experiment 2. The equations correspond to those presented in Table~\ref{tab:exp2}. The y-axis represents the \textit{Intervallu} and x-axis represents the \textit{Anomalia coaequata}. The true values for the \textit{Intervallu} from the Rudolphine tables are also plotted and labeled "original".}
    \label{fig:exp2}
\end{figure}
\begin{table}[]
    \centering
    \begin{tabular}{|c|c|c|}
    \hline
    Eqn No. & Eqn & MSE \\
    \hline
    1 & $0$ & 2.33 \\
    2 & $1.50000000000000$ & 0.0106 \\
    3 & $pi/2$ & 0.0123 \\
    4 & $1.65306122448980$ & 0.0268 \\
    5 & $1.66666666666667 - 0.09\times x_0^2$ & 0.048 \\
    6 & $(2.78 - 0.26\times x_0^2)^0.5$ & 0.0767 \\
    7 & $acos(-0.02\times x_0^3 + 0.09\times x_0^2 - 0.1)$ & 0.000169 \\
    8 & $1/(-0.01\times x_0^3 + 0.04\times x_0^2 + 0.6)$ & 0.000706 \\
    9 & $(0.01\times x_0^3 - 0.04\times x_0^2 + 1.29)^2$ & 0.000458 \\
    10 & $acos(-0.02\times x_0^3 + 0.09\times x_0^2 + 0.01\times x_0 - 0.1)$ & 4.7e-05 \\
    11 & $(0.06\times x_0^3 - 0.26\times x_0^2 - 0.05\times x_0 + 2.78)^0.5$ & 5.61e-06 \\
    12 & $0.02\times x_0^3 - 0.09\times x_0^2 - 0.01\times x_0 + 1.67$ & 4.41e-05 \\
    \hline
    \end{tabular}
    \caption{Experiment 3 Results}
    \label{tab:my_label3}
\end{table}
\begin{figure}[h]
    \centering
    \includegraphics[width = 0.7\textwidth]{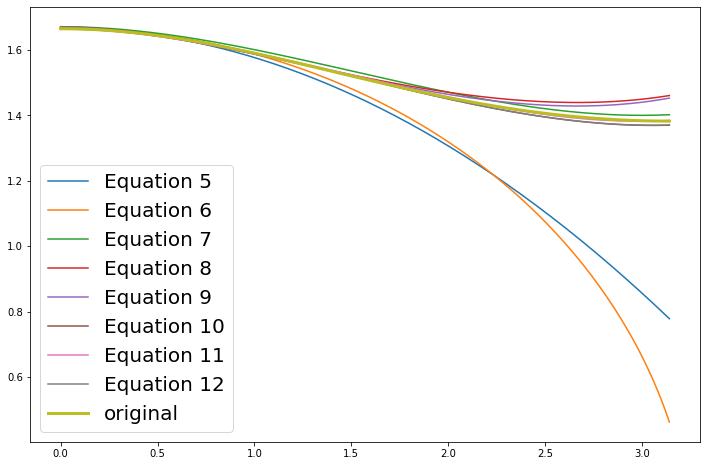}
    \caption{A plot of the results of Experiment 3. The equations correspond to those presented in Table~\ref{tab:exp3}. The y-axis represents the \textit{Intervallu} and x-axis represents the \textit{Anomalia coaequata}. The true values for the \textit{Intervallu} from the Rudolphine tables are also plotted and labeled "original".}
    \label{fig:exp3}
\end{figure}
\begin{table}[]
    \centering
    \begin{tabular}{|c|c|c|}
    \hline
    Eqn No. & Eqn & MSE \\
    \hline
    1 & $0$ & 2.33 \\
    2 & $1.50000000000000$ & 0.0106 \\
    3 & $1.56250000000000$ & 0.0115 \\
    4 & $0.142857142857143\times x_0 + 1.5$ & 0.000309 \\
    5 & $x_0/(x_1 + 6) + 1.5$ & 0.000416 \\
    6 & $1/(0.666666666666667 - 0.0557172402393568\times x_0)$ & 0.000265 \\
    7 & $1.510957104465+(x_0/((((((x_1+1)+1)+1)+1)+1)+1))$ & 0.000179 \\
    8 & $1/(0.662428081035614 - 0.0612907484173775\times x_0)$ & 7.75e-07 \\
    9 & $0.140863761305809\times x_0 - 0.0146051803603768\times x_1 + 1.52623379230499$ & 1.1e-06 \\
    10 & $(0.662428081035614 - 0.0612907484173775\times x_0)^(-1.001335978508)$ & 6.01e-10 \\
    \hline
    \end{tabular}
    \caption{Experiment 4 Results}
    \label{tab:my_label4}
\end{table}
\begin{figure}[h]
    \centering
    \includegraphics[width = 0.7\textwidth]{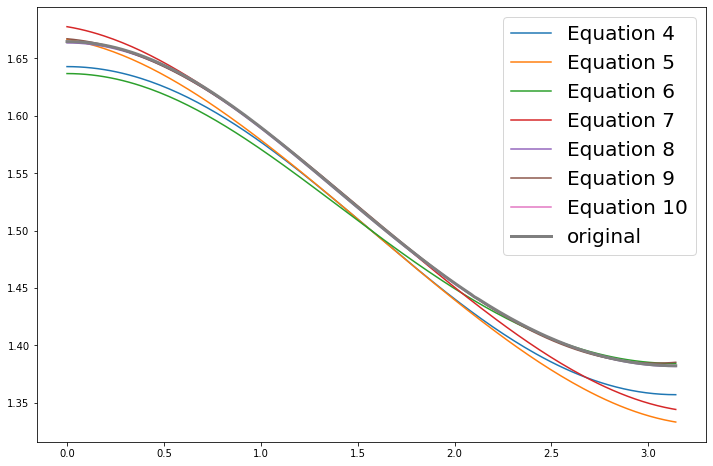}
    \caption{A plot of the results of Experiment 4. The equations correspond to those presented in Table~\ref{tab:exp4}. The y-axis represents the \textit{Intervallu} and x-axis represents the \textit{Anomalia coaequata}. The true values for the \textit{Intervallu} from the Rudolphine tables are also plotted and labeled "original".}
    \label{fig:exp4}
\end{figure}
\end{document}